%% file: main.tex
\documentclass[runningheads]{llncs}

\usepackage{eccv}

\usepackage{eccvabbrv}

\usepackage{multirow}
\usepackage{times}
\usepackage{epsfig}
\usepackage{graphicx}
\usepackage{amsmath}
\usepackage{amssymb}

\usepackage{booktabs}
\usepackage{epigraph}
\usepackage{enumitem}
\usepackage{float}
\usepackage{chngcntr}
\usepackage{soul}
\usepackage{caption}
\usepackage{subcaption}
\usepackage{setspace} 
\usepackage{zref-xr}
\usepackage{colortbl}
\usepackage[accsupp]{axessibility}  
\definecolor{baselinecolor}{gray}{.8}

\usepackage[pagebackref,breaklinks,colorlinks,citecolor=eccvblue]{hyperref}

\usepackage{orcidlink}
\input{preamble}
\usepackage{colortbl}
\usepackage{algorithm}
\usepackage{algorithmic}
\newcommand{\ours}{DITO\xspace}
\newcommand{\detection}{region-centric\xspace}
\newcommand{\Detection}{Region-centric\xspace}
\newcommand{\gray}[1]{\textcolor{gray}{{#1}}}

\newcommand{\blue}[1]{\color[HTML]{3166FF}{{#1}}}
\newcommand{\red}[1]{\color[HTML]{C41E3A}{{#1}}}
\newcommand{\white}[1]{\color[HTML]{FFFFFF}{{#1}}}

\begin{document}

\title{\Detection Image-Language Pretraining for Open-Vocabulary Detection} 


\author{Dahun Kim \quad
Anelia Angelova \quad
Weicheng Kuo}

\authorrunning{D.~Kim et al.}

\institute{Google DeepMind}

\maketitle
\input{sections/0_abstract}    
\input{sections/1_intro}
\input{sections/2_related}
\input{sections/3_method}
\input{sections/4_experiment}

\input{sections/5_conclusion}

%
%
\bibliographystyle{splncs04}
\bibliography{main}
\clearpage
\input{sections/6_appendix}
\end{document}

%% file: preamble.tex
\DeclareRobustCommand\onedot{\futurelet\@let@token\@onedot}
\def\eg{\emph{e.g. }} 
\def\ie{\emph{i.e. }} 
\def\etal{\emph{et al. }}

\usepackage{url}
\usepackage{graphicx}

\newcommand{\figref}[1]{Fig.~\ref{#1}}
\newcommand{\tabref}[1]{Table~\ref{#1}}
\newcommand{\eqnref}[1]{Eqn.~(\ref{#1})}
\newcommand{\secref}[1]{Sec.~\ref{#1}}
\renewcommand{\paragraph}[1]{\vspace{1mm}\noindent\textbf{#1}}
\usepackage{multirow, makecell}
\definecolor{baselinecolor}{gray}{.9}
\newcommand{\baseline}[1]{\cellcolor{baselinecolor}{#1}}
\usepackage{xcolor, soul}
\sethlcolor{baselinecolor}

\usepackage{pifont}
\usepackage{colortbl}
\usepackage{xspace}
\usepackage{tabu}
\usepackage{amssymb}
\newlength\savewidth
\newcommand{\tablestyle}[2]{\setlength{\tabcolsep}{#1}\renewcommand{\arraystretch}{#2}\centering\footnotesize}

\makeatletter
\newcommand{\thickhline}{%
    \noalign {\ifnum 0=`}\fi \hrule height 1pt
    \futurelet \reserved@a \@xhline
}
\makeatother

%% file: sections/0_abstract.tex
\begin{abstract}
We present a new open-vocabulary detection approach based on region-centric image-language pretraining to bridge the gap between image-level pretraining and open-vocabulary object detection. At the pretraining phase, we incorporate the detector architecture on top of the classification backbone, which better serves the region-level recognition needs of detection by enabling the detector heads to learn from large-scale image-text pairs. Using only standard contrastive loss and no pseudo-labeling, our approach is a simple yet effective extension of the contrastive learning method to learn emergent object-semantic cues. In addition, we propose a shifted-window learning approach upon window attention to make the backbone representation more robust, translation-invariant, and less biased by the window pattern. On the popular LVIS open-vocabulary detection benchmark, our approach sets a new state of the art of 37.6 mask AP$_r$ using the common ViT-L backbone and public LAION dataset, and 40.5 mask AP$_r$ using the DataComp-1B dataset, significantly outperforming the best existing approach by +3.7 mask AP$_r$ at system level. On the COCO benchmark, we achieve very competitive 39.6 novel AP without pseudo labeling or weak supervision. In addition, we evaluate our approach on the transfer detection setup, where it demonstrates notable improvement over the baseline. Visualization reveals emerging object locality from the pretraining recipes compared to the baseline.\footnote{project page: \href{https://github.com/google-research/google-research/tree/master/fvlm/dito}{github.com/google-research/google-research/tree/master/fvlm/dito}}
\end{abstract}

%% file: sections/1_intro.tex
\vspace{-7mm}
\section{Introduction}
\vspace{-2mm}

To understand and localize all objects and entities in the visual world has been a foundational problem in computer vision and machine learning. This capability unlocks a broad array of compelling applications from self-driving cars to search and recommendation. However, existing object detectors typically rely on human-annotated regions and class labels. These annotations are costly and unscalable in terms of the number of categories \eg O(1K) and the number of images \eg O(100K).

\begin{figure}
\begin{centering}
\includegraphics[width=\linewidth]{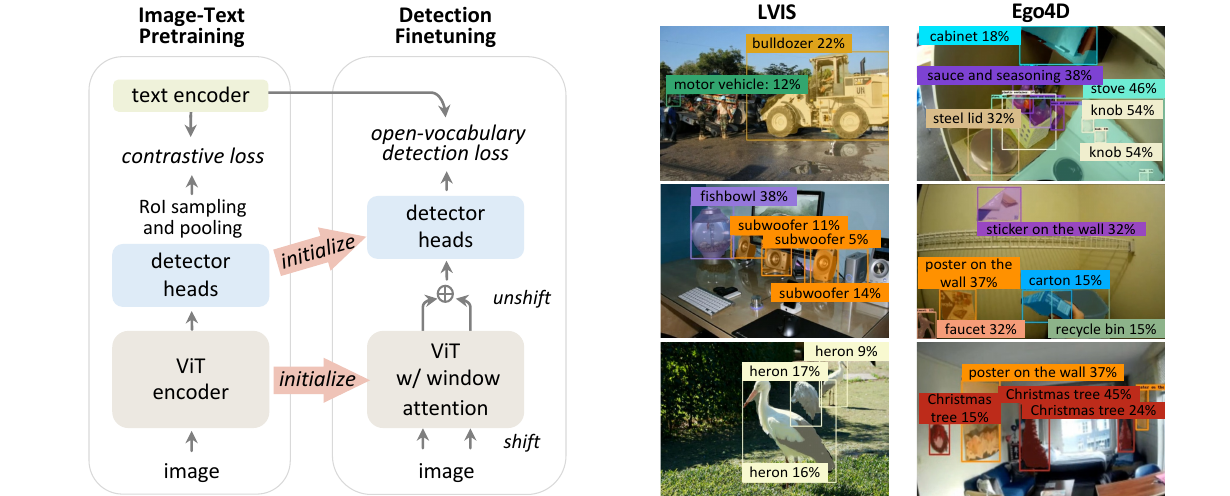}
\caption{\textbf{\ours overview (left):} Existing image-text pretraining methods for open-vocabulary detection~\cite{rovit,kim2023contrastive} only update the vision transformer backbone and finetune detector heads from scratch.
We propose to pretrain both backbone and detector heads directly from the large-scale image-text paired data, without a need for pseudo labeling or box annotations. In open-vocabulary detection finetuning, we introduce a simple shifted-window learning method to produce more robust representations from the pretrained vision transformer. 
\textbf{\ours prediction (right):} LVIS results only show the novel categories (\eg, \textit{bulldozer, fishbowl, subwoofer, heron}). While Ego4D is a real-world and out-of-distribution data, many unseen objects are detected (\eg, \textit{steel lid, sticker on the wall, recycle bin}). Best viewed with zoom in.
}
\vspace{-4mm}
\label{fig:teaser}
\end{centering}
\end{figure}

The open-vocabulary detection (OVD) task~\cite{Zareian_2021_CVPR} has been introduced to overcome both limitations by pretraining on larger-scale image-text data before finetuning for detection tasks. In particular, recent open-vocabulary detection approaches are mostly based on Contrastive Language-Image Pretraining (CLIP)~\cite{radford2021clip}, representing each category as a text embedding rather than a discrete label. This enables the detectors to localize objects based on any user-provided text queries unavailable during training.

Most existing open-vocabulary detection works assume the pretrained CLIP backbone is given, and focus on techniques such as knowledge distillation~\cite{gu2022openvocabulary,du2022learning}, weak supervision~\cite{zhou2022detecting}, pseudo labeling~\cite{zhong2021regionclip,rasheed2022bridging,zhao2022exploiting,huynh2022open}, and frozen backbone application~\cite{kuo2022fvlm}, using the pretrained backbone.
Consequently, during detection finetuning, the detector heads often need to be trained from scratch.
This tends to result in sub-optimal generalization because the detector heads are trained on the limited vocabulary of detection datasets, while only the backbone contains the knowledge of open-vocabulary concepts.

Several studies have integrated detection models into CLIP pretraining. For instance, RegionCLIP~\cite{zhong2021regionclip} employs an off-the-shelf detector during CLIP pretraining to obtain proposals on the image-text data and subsequently generate pseudo region-text labels. However, their pretraining only updates the image backbone, with the detector heads exclusively trained on detection data. These pseudo labeling-based pretraining methods~\cite{zhong2021regionclip,minderer2023scaling} require multi-stage training and handcrafted processing to generate high-quality pseudo labels, and results in increased complexity and cost of training. Similarly, other approaches such as GLIP~\cite{li2021grounded,zhang2022glipv2}, Grounding DINO~\cite{liu2023grounding}, CoDet~\cite{ma2024codet} and DetCLIP~\cite{detclip,yao2023detclipv2} integrate detector architectures in CLIP training.
However, their joint training requires additional detection and visual grounding datasets and complex multitask learning setups.

In this paper, we propose a simple yet effective solution, Region-centric Pretraining approach, which incorporates detector modules into CLIP pretraining without the need for pseudo labeling or box annotations. This method involves generating random box regions across feature pyramid levels, followed by feature pooling over these regions. Subsequently, an image-text contrastive loss is applied, encouraging text-aligned region features to contribute more to the whole image representation. 
Our approach not only facilitates the warm-starting of detector heads in finetuning, but also leads to emergent representations with more localized semantic information compared to the baseline CLIP backbone, as demonstrated in our experiments.
Compared to pseudo-labeling techniques~\cite{zhong2021regionclip,zhou2022detecting,wu2023aligning,zhong2021dap}, our approach can be viewed as an extension of contrastive models to bypass the need for offline object proposal generation, which could be cumbersome for large-scale image-text data.

In addition, we focus on the Vision Transformer (ViT) based CLIP models.
While CLIP ViTs excel in zero-shot recognition, their direct integration to open-vocabulary detector has gained less traction compared to their ConvNet counterparts~\cite{zhong2021regionclip,kuo2022fvlm,wu2023cora}. 
The typical pretraining of CLIP ViT backbones on lower resolutions, followed by adaptation to higher resolution detection images, presents challenges due to increased computational demands and the risks of breaking locality structure in pretrained features. Although the windowed attention technique of ViTDet~\cite{li2022exploring,rovit} can reduce computation and preserve locality structure, it introduces bias from the fixed window patterns at the same time. To address this, we propose Shifted-Window Learning (SWL) approach to enhance information mixing across fixed windows and mitigate grid pattern bias. Unlike the Swin Transformer~\cite{swin}, which applies shifted window layer by layer, SWL applies shifted windowed attention as a separate forward pass using the same ViT backbone. This simple strategy enhances the windowed attention representation when applied low-res trained ViT backbone to high-res images, ensuring compatibility with vanilla ViT backbones pretrained without shifted windows.
Apart from improving windowed attention, SWL better preserves the open-vocabulary knowledge of pretrained features compared to full-attention ViT perhaps due to its emphasis on local cues.

Incorporating both region-centric pretraining and shifted-window learning, our approach is called \ours (Detection-aligned Image-Text pretraining for Open-vocabulary detection). \ours serves to narrow the gap between image-text pretraining and open-vocabulary detection, and obtain better generalization.
The best \ours model achieves 37.6 mask AP$_r$ on the widely used LVIS open-vocabulary detection benchmark, surpassing the previous best approach by +3.7 AP$_r$. It achieves the state-of-the-art 40.5 mask AP$_r$ when pretrained on the DataComp-1B dataset. On the COCO benchmark, \ours achieves a very competitive 39.6 novel AP without using pseudo-labels or joint training.
In summary, our contributions are:
\begin{itemize}
\item We present a novel region-centric pretraining approach for open-vocabulary detection by integrating detector heads on top of the image backbone into CLIP pretraining. This learns better detection-oriented features from large-scale image-text data without a need for pseudo labeling or extra box annotations.
\item We propose the Shifted-Window Learning technique to produce more robust and translation-invariant representation from pretrained CLIP ViT for open-vocabulary detection. 
\item Our approach significantly outperforms the state-of-the-art methods on LVIS open-vocabulary detection benchmark, including larger models and pseudo labeling-based approaches, and achieves very competitive performance on COCO benchmark and transfer detection to Objects365.
\end{itemize}

%% file: sections/2_related.tex
\vspace{-4mm}
\section{Related Work}
\vspace{-2mm}
\paragraph{Open-vocabulary detection.}\quad
Conventional closed-set object detectors exhibit great performance but are limited in their vocabulary size. Motivated by the strong zero-shot abilities of Vision-Language Models (VLMs) like CLIP~\cite{radford2021clip}, open-vocabulary detection has shown notable progress in recent years. 
Efforts have been directed towards leveraging pretrained CLIP models for detection, with various techniques such as knowledge distillation~\cite{gu2022openvocabulary} and prompt optimization~\cite{du2022learning}. Also, there are works that directly utilize the pretrained CLIP backbone by adding new detection heads either by setting the backbone frozen~\cite{kuo2022fvlm,wu2023cora} or finetunable~\cite{minderer2022simple,rovit}. Many top-performing approaches~\cite{zhong2021regionclip,li2021grounded,pb-ovd,feng2022promptdet,minderer2023scaling} rely on pseudo labeling techniques which aim to mitigate the issue of catastrophic forgetting in detection finetuning. However, these self-training methods often necessitate multi-stage detection training~\cite{zhong2021regionclip,wu2023aligning}, extra steps for generating high-quality pseudo labels~\cite{minderer2023scaling}, or the use of off-the-shelf detector modules~\cite{zhong2021regionclip,rasheed2022bridging}.
Unlike these approaches, our method focuses on both upstream image-text pertraining and downstream open-vocabulary detection without the need for pseudo labeling techniques. 
We demonstrate significant improvements in open-vocabulary detection.

\paragraph{Region-aligned Vision-Language Models.}\quad
Driven by the progress in aligning image-text representations~\cite{radford2021clip, schuhmann2021laion}, several studies have aimed to integrate region-level alignment into CLIP pretraining. For instance, RegionCLIP~\cite{zhong2021regionclip} learns region-word alignment by generating pseudo region-text pairs. However, their pretraining requires an off-the-shelf detector for the pseudo labeling, and it solely focuses on training the image backbone, relying on the off-the-shelf detector during inference. Other approaches like GLIP~\cite{li2021grounded,zhang2022glipv2}, Grounding DINO~\cite{liu2023grounding}, DetCLIP~\cite{detclip,yao2023detclipv2}, CoDet~\cite{ma2024codet} integrate detector architectures in CLIP training to explicitly align regions with words. However, they require additional detection or visual grounding annotations for explicit region-level supervision, resulting in complex multitask learning. To our knowledge, our approach is the first work that incorporates detector modules in image-text pretraining without relying on any box annotations.

\paragraph{Vision Transformers in open-vocabulary detection.}\quad
Regarding backbone architecture, ConvNet, ViT or hybrid models~\cite{swin} have been used in open-vocabulary detection. While ViT-based CLIP models exhibit superior capability in zero-shot recognition, their adaptation to open-vocabulary detection is relatively less explored compared to ConvNet-based CLIP models. A notable example is OWL-ViT~\cite{minderer2022simple, minderer2023scaling}, which finetunes the pretrained ViT on higher-resolution detection images while maintaining fully global attentions. However, employing full attention models can be computationally demanding for large images. RO-ViT~\cite{rovit} proposes region-aware positional embeddings that aid in the generalization of CLIP ViT onto detection finetuning. These methods~\cite{rovit,kim2023contrastive} adopt windowed attention from ViTDet~\cite{li2022exploring} for adaptation to detection. In this paper, to further enhance information mixing across fixed windows while preserving the locality structure of the pretrained lower-resolution features, we propose shifted-window learning to mitigate the window-induced bias and improve open-vocabulary detection.

\paragraph{Self-supervised pretraining for visual tasks.}\quad
Self-supervised learning has emerged as a promising paradigm to learn object features for complex visual tasks such as detection, given the challenge of scaling up human annotations.
Most relevant direction is contrastive learning, where the contrastive samples can take the forms of augmented images~\cite{chen2020simple}, sliding windows~\cite{Xiao_2021_ICCV}, object proposals~\cite{SoCo}, or point samples~\cite{Bai_2022_CVPR}. Some alternative strategies like pseudo-labeling~\cite{zhong2021dap} and pixel reconstruction~\cite{he2022masked} have also proven effective. While the majority of these methods have focused on learning from images without textual context, and applying to closed-vocabulary detection, \ours leverages large image-text data to tackle the more demanding open-vocabulary detection task, without a need for offline proposal generation~\cite{SoCo,zhong2021dap}.

%% file: sections/3_method.tex
\vspace{-3mm}
\section{Method}
\vspace{-2mm}

We address the problem of open-vocabulary object detection. At training time, the model can access the class and box labels of base categories ($C_B$). At test time, the model is tasked with detecting objects from a set of novel categories ($C_N$) not present in the training set. 
To achieve this, we leverage pretrained vision and language models (VLMs) as in prior studies~\cite{rovit,gu2022openvocabulary,zhong2021regionclip,kuo2022fvlm}. 

More specifically, we leverage recent advances in ViT-based detectors~\cite{li2022exploring,rovit} for their promising results. However, instead of solely taking the pretrained ViT backbone, we demonstrate how to enhance the VLMs by integrating detector heads into the CLIP pretraining process, referred to as Region-centric image-text Pretraining (REP). Additionally, for detection finetuning, we propose a Shifted-Window Learning (SWL) strategy to enhance the adaptation of the pretrained ViT model to open-vocabulary detection task. By combining these approaches, \ours achieves significant improvements in open-vocabulary detection over prior arts.

\begin{figure*}[t]
\centering
\includegraphics[width=\linewidth]{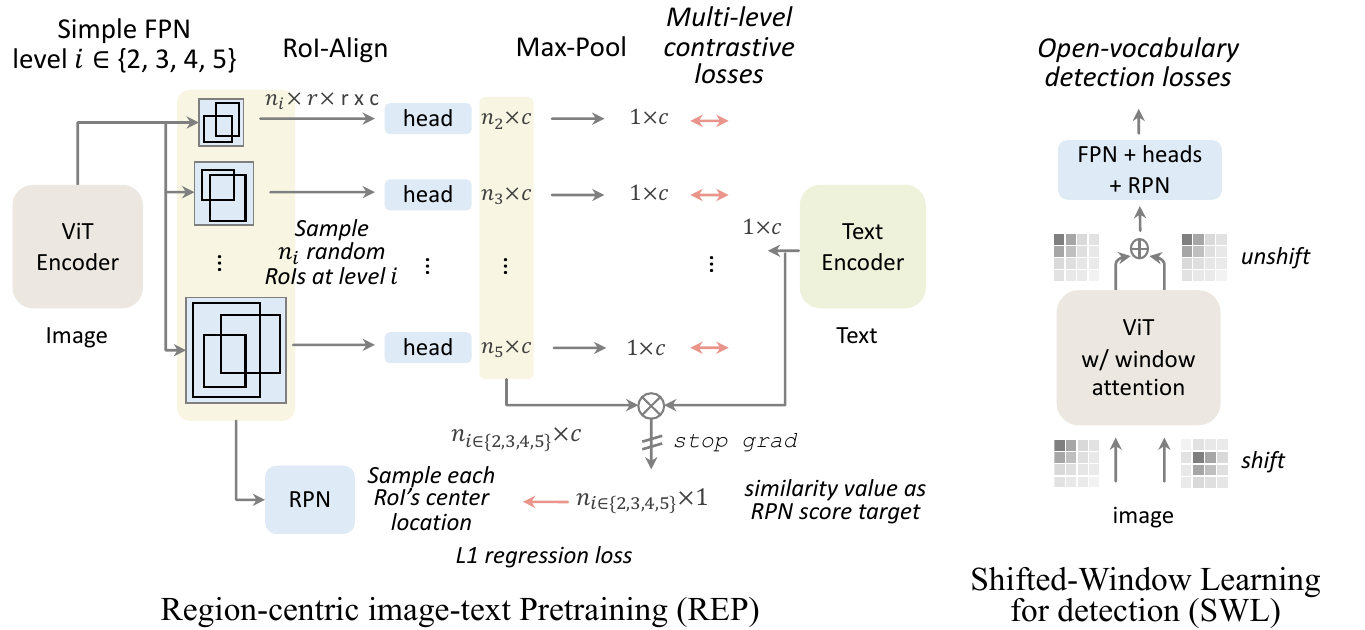}
\vspace{-4mm}
\caption{\textbf{\ours method.} \Detection image-text pretraining (left): We train the detector heads (e.g. FPN~\cite{li2022exploring,lin2017feature}, Faster RCNN head~\cite{ren2015faster}, and RPN~\cite{ren2015faster}) upon a ViT encoder backbone with multi-level image-text contrastive loss to bridge the gap between image-text pretraining and open-vocabulary detection. Shifted-window learning for detection (right): We roll the image and combines the shifted features with the original features to mitigate the bias of windowed attention grid~\cite{li2022exploring}, and produce more robust semantic representation.
}
\vspace{-5mm}
\label{fig:overview}
\end{figure*}

\vspace{-2mm}
\subsection{Preliminaries}
\vspace{-2mm}

\label{sec:preliminaries}

\paragraph{Baseline.}\quad
Our baseline approach is RO-ViT~\cite{rovit}, a state-of-the-art ViT-based method for open-vocabulary detection. RO-ViT introduces CLIP pretraining with a new positional embedding scheme called cropped positional embedding (CPE). CPE involves randomly cropping and resizing the standard whole-image positional embedding during pretraining to enhance the ViT's generalization onto region-level recognition task and higher resolution detection inputs downstream.
For detection finetuning, it adopts ViTDet~\cite{li2022exploring} architecture, initialized with the pretrained ViT backbone. In the following, we describe the image-text pretraining and the downstream open-vocabulary detection.

\paragraph{Image-text pretraining.}\quad
We adopt dual-encoder CLIP pretraining widely used in existing works~\cite{radford2021clip,schuhmann2021laion}.
The image embeddings $\{{v}\}$ and text embeddings $\{{l}\}$ are the average-pooled outputs from the image and text encoders, respectively. As in previous works, we compute the dot product of the embeddings in batch $B$, and scale it by a learnable temperature $\tau$ before applying the InfoNCE loss~\cite{oord2018representation,radford2021clip}. Mathematically, the image-to-text (I2T) loss can be expressed as: 
\begin{equation}\label{eqn:contrastive}
L_{\text{I2T}} = -{1 \over {B}} \sum_{i=1}^{B} \log({\text{exp}(v_{i}l_{i} / \tau) \over { \sum_{j=1}^{B} \text{exp}(v_{i} l_{j} / \tau)  }}).
\end{equation}
The text-to-image (T2I) loss is symmetrical by exchanging the inner/outer summation loops. The total contrastive loss $L_{con}$ is obtained by $L_{con} = (L_{\text{I2T}} + L_{\text{T2I}}) / 2$. As mentioned above, we adopt the cropped positional embeddings (CPE) following~\cite{rovit}.

\paragraph{Open-vocabulary detection finetuning.}\quad
At the fine-tuning stage, our detection finetuning recipe follows previous studies~\cite{Zareian_2021_CVPR,gu2022openvocabulary, kuo2022fvlm, rovit}. 
During the training phase, we use the RoI-Align~\cite{he2017mask} feature as the detection embedding for each detected region. We replace the fixed-size classifier layer with the text embeddings of base categories ($C_B$). The detection score $p_i$ is determined by calculating the cosine similarity between the region embedding $r_i$ and text embeddings of base categories ($C_B$) followed by a softmax operation. We prepend an additional background class embedding to $C_B$ and use the term ``background" to represent the background category. Any proposals that do not match to any base category annotations are treated as background during training. It is important that the text embeddings are computed from the same text encoder as from the image-text pretraining. During testing, we expand the text embeddings to include the novel categories ($C_B \cup C_N$), resulting in ($C_B \cup C_N$ + 1) categories including the background. We calculate the detection scores ($p_i$) as the cosine similarity between the region embeddings ($r_i$) and the expanded text embeddings. Apart from the detection embedding ($r_i$), we extract the VLM embedding~\cite{kuo2022fvlm} of region $i$ by RoI-Align at the last ViT backbone feature map. The VLM score ($z_i$) is calculated as the cosine similarity with the text embeddings of the combined categories ($C_B \cup C_N$).

To compute the open-vocabulary detection score (${s_i}^{\text{ens}}$), we ensemble the detection and VLM scores by geometric means~\cite{gu2022openvocabulary,kuo2022fvlm}. The formula is as follows:
\begin{equation}\label{eqn:combine-score}
{s_i}^{\text{ens}} = \begin{cases}
    z_i^{(1-\alpha)} \cdot p_i^\alpha & \text{if } i \in C_B\\
    z_i^{(1-\beta)} \cdot p_i^\beta & \text{if } i \in C_N
\end{cases}
\end{equation}
Here, $\alpha, \beta$ are floats $\in$ $[0, 1]$  that control the weighting of base versus novel categories. The background score comes from the detection score  ($p_i$) alone, because we observe the VLM score of ``background" class is often  less reliable.

\vspace{-2mm}
\subsection{\Detection Image-Text Pretraining}
\vspace{-0mm}

\label{sec:rep}
Standard image-text pretraining uses classification architectures (\eg ViT backbone followed by global pooling) as the language supervision occurs at the image level rather than the region level. 
Subsequently, for downstream detection, new detection heads are introduced and trained from scratch on a limited set of detection categories~\cite{kuo2022fvlm,rovit,kim2023contrastive}. 
To fully utilize the knowledge embedded in large-scale image-text data, we propose \Detection Pretraining (REP) which integrates the detector modules during the CLIP pretraining phase. Specifically, given access to image-text paired data but lacking box labels, our pretraining focuses on the \textit{region-recognition pathway} of a detector, encompassing components like the backbone, FPN~\cite{lin2017feature,li2022exploring}, RoI-Align~\cite{he2017mask}, RPN-objectness~\cite{ren2015faster}, and Faster RCNN-classifier~\cite{ren2015faster}. Consequently, the detector heads can be warm-started from the knowledge of large image-text data, thereby improving the generalization capability. To our knowledge, we are the first to integrate detector modules in image-text pretraining without box labels, and our experiments demonstrate clear benefits of our approach in open-vocabulary detection.

\paragraph{Detector head learning from random regions.}\quad 
\figref{fig:overview} (left) illustrates our region-centric pretraining system.
Following existing works~\cite{rovit,kim2023contrastive}, we adopt SimpleFPN~\cite{li2022exploring} and Faster R-CNN~\cite{ren2015faster} models to remove the detector differences and study the benefits of our \detection pretraining. Specifically, the multi-level feature pyramid is computed from the ViT backbone. Then, RoI-Align~\cite{he2017mask} and Faster R-CNN classifier head are applied to these feature maps to match the classification pathway in pretraining with the region-recognition pathway in detection finetuning (see~\tabref{tab:ablation:dito_pretraining} for ablations).

For each level $i$ of the feature pyramid, we randomly generate $n_i$ box regions uniformly over the image by sampling the box size $h,w \thicksim \textit{Uniform}(0.2, 0.5)$ and aspect ratio $h/w \thicksim \textit{Uniform}(0.5, 2.0)$. The $n_i$ value is set proportional to the size of the $i$-th feature map so that larger feature map would be covered by more regions. We extract the RoI-features of each region by RoI-Align, and feed them through the region classifier head~\cite{ren2015faster} to obtain the RoI embeddings.

\paragraph{Multi-level image-text supervision.}\quad After computing the RoI embeddings across pyramid levels, we perform a max pooling  over the RoI embeddings per-level to obtain an image embedding for each pyramid level. Intuitively, max pooling allows the representation to focus on salient regions and discriminative features, thereby learning region-level information without explicit supervision. Then we apply the standard image-text contrastive loss (see ~\eqnref{eqn:contrastive}) on each feature level separately, which aids the learning of rich semantic and object knowledge within every feature map (see \tabref{tab:ablation:dito_roi} for ablations). The losses from all levels are weighted equally and summed together. 
Without explicit region-level supervision, the max pooling over regions encourages the more salient, text-aligned region features to contribute more to the whole image representation in the contrastive loss.
Our experiments show emergent region-text alignments from the multi-level training, where the feature maps possess more localized semantic information compared to the baseline CLIP~\cite{rovit} backbone (see \figref{fig:heatmap}).

Different from pseudo-labeling techniques~\cite{zhong2021regionclip,feng2022promptdet,huynh2022open, SoCo, wu2023aligning} that require additional steps to generate and store annotations, our approach learns the detector heads on the fly without a need to compute or store object proposals.

\paragraph{Region proposal network learning.}\quad
\figref{fig:heatmap} shows that the the learnt multi-level representations exhibit localized semantics well-aligned with the text query, \ie the salient regions have higher similarity with the text relative to the background. Motivated by this observation, we employ the multi-level visual-text similarity as a supervisory signal for training the Region Proposal Network (RPN)~\cite{ren2015faster}. Specifically, we compute the cosine similarity between each RoI embedding and the text, and use it as the target RPN score for the center location of each RoI. Any negative dot product value is mapped to zero to keep the target score in range $[0, 1]$.  We use L1 regression loss and set the loss weight equal to the multi-level contrastive loss (see Sec. A in supp.) The losses are propagated only through the RoI centers and other pixels are ignored. 
Note that the box regression of the RPN is \textit{not} trained here but learnt later through the detection finetuning, as we only use the image-text paired data without any box annotations.

\vspace{-2mm}
\subsection{Shifted-Window Learning for Detection}
\vspace{-2mm}
\label{sec:swl}
The CLIP ViT backbones are typically pretrained on lower resolutions (\eg 224$\times$224) and then adapted to higher resolution detection images (\eg 1024$\times$1024). While the detection task benefits from global information, directly applying the pretrained ViT on high-resolution inputs is not only computationally intensive but can also compromise the preservation of the locality structure of the pretrained lower-resolution features.

The adaptation such as windowed attention in ViTDet~\cite{li2022exploring} effectively reduces the computation and is thus also adopted by our baseline RO-ViT~\cite{rovit} detector. However, we observed that the backbone representation is still biased by the fixed-size grid pattern of the windowed attention, compromising the representation power of the pretrained ViT. To improve information mixing across the fixed windows and mitigate the bias of the grid pattern, we propose the Shifted-Window Learning (SWL) approach.

\paragraph{Network architecture.}\quad \figref{fig:overview} (right) and Algorithm \ref{alg:swl} describe the SWL algorithm. 
The standard ViT consists of a patchifying layer and a set of transformer layers. After feeding the image through the patchifying layer, we obtain a feature map $x$ of shape $(h, w, d)$. 
This feature map $x$ is fed through the rest of the ViT with windowed attention layers on a grid $K\times K$, and $L$ global attention layers evenly spaced throughout the ViT (where $L$ = 4) following~\cite{li2022exploring}, resulting in output $y$. 
In parallel, we create another copy of $x$, which is rolled along both $h$ and $w$ axes by $s$ pixels. The elements that roll beyond the last position are reintroduced from the first. We carefully design the attention masks such that the rolled around patches would not attend to the patches on the other side of the image (see the right figure of Algorithm~\ref{alg:swl}). The shift size $s$ is set as the half of the attention window size $M$ (\ie $s = M//2$). 
Empirically, the window size $M$ = 16 equals the image size (\eg 1024) divided by the product of patch size (\eg $P$ = 16) and the grid size (\eg $K$ = 4). The shifted feature map $x'$ is then processed through the rest of the ViT in the same manner, resulting in output $y'$ in the same shape $(h, w, d)$ as $y$. We then unshift $y'$ and combine it with $y$ by averaging. We apply the above shifted window operations in detection finetuning and inference times.

\paragraph{Comparison with Swin Transformer.}\quad Compared to the Swin Transformer~\cite{swin},  we apply the shifted-window ideas as separate forward passes, while Swin Transformer applies similar ideas in an alternating manner layer by layer. Our approach requires no change to the vanilla transformer architecture and is compatible with any ViT backbones pretrained without shifted windows (\eg \cite{radford2021clip}), whereas Swin Transformer requires specialized pretraining on the same architecture. Compared to the full-attention ViT~\cite{dosovitskiy2020image}, we observe that shifted-window ViT taps more effectively into the semantic knowledge of pretrained backbone than full global attention, perhaps because the window helps the model focus more on local cues and ignore the noises farther away.

{
\begin{algorithm}[t]
{
\begin{minipage}[b]{0.6\textwidth}
\centering
\scriptsize{\captionof{algorithm}{Shifted Window backbone}
\label{alg:swl}
\begin{algorithmic}
    \STATE $x$: image patch tokens + positional embeddings.\textcolor{gray}{~~~~~~~~~~~~~~~\# [h, w, d]}
    \STATE M: attention window size.
    \STATE s = M // 2
    \STATE $y$ = \texttt{forward\_vit\_with\_win\_attn}($x$, $M$)
    \STATE $x'$ = \texttt{np.roll}($x$, \texttt{shift=}[$s$, $s$], \texttt{axis=}[$0$, $1$])\textcolor{gray}{~~~~~~~~~~~~~~~~~~~\# shift}
    \STATE $y'$ = \texttt{forward\_vit\_with\_win\_attn}($x'$, $M$)
    \STATE $y''$ = \texttt{np.roll}($y'$, \texttt{shift=}[-$s$, -$s$], \texttt{axis=}[$0$, $1$])\textcolor{gray}{~~~~~~~~~~~\# unshift}
    \STATE \textbf{return} ($y$ + $y''$) / 2  \textcolor{gray}{~~~~~~~~~~~~~~~~~~~~~~~~~~~~~~~~~~~~~\# SWL backbone output}
\end{algorithmic}}
\end{minipage}
\begin{minipage}[b]{0.39\textwidth}
\centering
\includegraphics[width=1.0\linewidth]{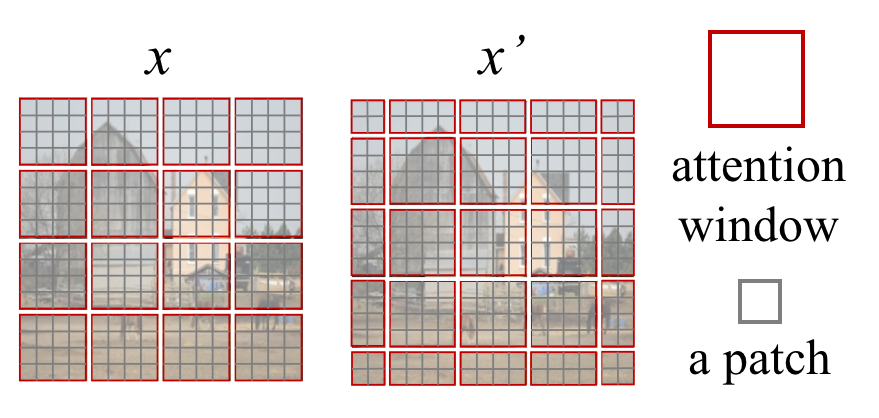}
\end{minipage}
}\vspace{-2mm}
\end{algorithm}
}

\vspace{-3mm}
\subsection{Distillation from Frozen ViT Backbone}
\vspace{-2mm}
\label{sec:dist}
While the ViT backbone adapts to the detection tasks, it may lose some of its pretrained open-vocabulary knowledge. Therefore, we propose a simple distillation approach which uses a separate frozen ViT to teach the finetuned ViT backbone during the detection finetuning. We use a cosine distance loss that aligns the RoI-Align embeddings extracted from the feature maps of both backbones. The cosine distance is computed for each RoI then averaged over all RoIs.
Empirically, we find it advantageous to add a 1$\times$1 Conv projection layer to the finetuned ViT backbone before the RoI-Align, allowing some flexibility in distillation when jointly trained with other detection objectives.
The auxiliary distillation loss is then added to other detection losses, with a loss weight $\gamma=1$ (see Sec. A in supp.) At inference, the ViT backbone features after the projection are used to compute the region VLM score $z_i$ (\secref{sec:preliminaries}). It is worth noting that the frozen ViT backbone is only used during training for distillation purposes and is removed at inference. 

While previous studies~\cite{gu2022openvocabulary,chen2023exploring} have utilized knowledge distillation from the CLIP models, their teacher CLIP models mostly operate on image crops in an offline process, thus needing multiple forward passes through the backbone for all RoIs. In contrast, our distillation operates efficiently on RoIs cropped from dense feature maps in a single forward pass during the detection finetuning with minimal overhead.

%% file: sections/4_experiment.tex
\vspace{-3mm}
\section{Experimental Results}
\vspace{-2mm}

\paragraph{Baseline reproduction.}\quad
As discussed in \secref{sec:preliminaries}, our baseline method is RO-ViT~\cite{rovit}, a leading ViT-based approach for open-vocabulary detection. RO-ViT introduces cropped positional embedding (CPE) in CLIP pretraining to enhance the generalization of the pretrained ViT onto the downstream detection. Additionally, it adopts global average pooling of the ViT features instead of CLS-token pooling, which simplifies the adaptation onto higher resolution inputs, and the extraction of region features (\eg RoI-Align) at the final layer.
We reproduce the CLIP pretraining of RO-ViT~\cite{rovit} using the widely adopted LAION-2B dataset~\cite{schuhmann2021laion}, as the OpenAI CLIP's WIT~\cite{radford2021clip} and ALIGN~\cite{align} datasets are not publicly available. We trained the CLIP models from scratch, following the same pretraining protocol and hyperparameters, including 500k iterations and 16k batch size (\ie 8B samples seen in training), 224$\times$224 image size, global average pooling, and cropped positional embedding. We use the standard InfoNCE loss (\eqnref{eqn:contrastive}) instead of their focal constrastive loss~\cite{rovit}. The following compares our reproduced CLIP with OpenAI CLIP~\cite{radford2021clip} and OpenCLIP~\cite{schuhmann2021laion}:

\begin{center}\vspace{-0.4em}
\tablestyle{4pt}{1.0}
\scriptsize{
\begin{tabular}{lcccc}
method & dataset   & {\# samples seen} & backbone  & {ImageNet Top-1 Acc.} \\
\midrule
CLIP~\cite{radford2021clip} & WIT-400M  & 13B   & ViT-L/14  & 75.5   \\
OpenCLIP~\cite{schuhmann2021laion} & LAION-2B               & 32B    & ViT-L/14  & 75.2 \\
RO-ViT CLIP~\cite{rovit} \textit{our repro.} & LAION-2B      & 8B    & ViT-L/16  & 73.9 \\
\end{tabular}\vspace{-0.4em}
}
\end{center}
On the popular zero-shot ImageNet classification benchmark, the LAION-2B pretraining matches or slightly underperforms the OpenAI CLIP with WIT-400M.

\paragraph{Pretraining setup.}\quad
After the above-mentioned baseline CLIP training, we apply our \detection pretraining (REP) where we freeze the image and text encoders trained in the first phase and introduce the detector heads. We use the Simple FPN~\cite{li2022exploring}, and classification layers of Faster R-CNN and RPN~\cite{ren2015faster}, where we replace the batch normalization with layer normalization. At the $i$-th pyramid level $i\in \{2, 3, 4, 5\}$, we randomly sample $n_i\in \{400, 200, 100, 50\}$ box regions and compute their RoI-Align features. We use a short training cycle of 30k iterations, 4k batch size, 256$\times$256 image size, AdamW optimizer with an initial learning rate of 1e-4 with linear decay. 
For both phases of image-text pretraining, we use the publicly available LAION-2B~\cite{schuhmann2021laion} dataset.

\paragraph{Detection finetuning setup.}\quad
As noted in \secref{sec:rep}, we adopt the ViTDet~\cite{li2022exploring} with SimpleFPN as our detector following the baseline works~\cite{rovit, kim2023contrastive}. We follow the same finetuning settings of~\cite{rovit}. Specifically, we train the detector with image size $1024\times1024$ and use windowed attention in the backbone with grid size 4$\times$4. The learning rate for the backbone is set lower as 0.6$\times$ of the detector head layers. 
We use $\alpha$=0.3, $\beta$=0.65 for score combination in \eqnref{eqn:combine-score}.
The text embedding of each category is calculated as the average over the CLIP prompt templates. We use the batch size 128, the SGD optimizer with momentum 0.9. The initial learning rate and iterations are set to 0.18 and 36.8k for LVIS, and 0.02 and 11.3k for COCO datasets.

\begin{table}[t]
\centering
\tablestyle{3.0pt}{0.9}
\scriptsize{
\begin{tabular}{lccccll}
\toprule
method & \makecell[c]{pretraining \\ model} & \makecell[c]{pretraining \\ data} & \makecell[c]{detector \\ backbone }  & \makecell[c]{w/ pseudo \\ box labels} & 
\makecell[l]{\bf{mask} \\ \bf{AP$_r$}}
& \makecell[l]{\gray{mask} \\ \gray{AP}}  
\\
\midrule
\bf\textit{ConvNet based:} & & & & \\
OV-DETR~\cite{zang2022open}            & ViT-B/32 & CLIP-400M & R-50  & -    & 17.4      & \gray{26.6} \\
Kaul \etal~\cite{kaul2023multi}   & ViT-B/32   & CLIP-400M & R-50     & - & 19.3 & \gray{30.6} \\
DetPro-Cascade~\cite{du2022learning}      & ViT-B/32  & CLIP-400M & R-50   & -        & 20.0      & \gray{27.0} \\
Rasheed~\cite{rasheed2022bridging}      & ViT-B/32  & CLIP-400M    & R-50    & -       & 21.1      & \gray{25.9}\\
BARON~\cite{wu2023aligning}   & ViT-B/32   & CLIP-400M & R-50     & - & 22.6 & \gray{27.6} \\
CoDet~\cite{ma2024codet}  & R-50 & CLIP-400M + CC3M     & R-50       & -    & 23.4      & \gray{30.7}\\
EdaDet~\cite{shi2023edadet}  & R-50 & CLIP-400M    & R-50       & - & 23.7      & \gray{27.5}\\
VL-PLM~\cite{zhao2022exploiting} & ViT-B/32  & CLIP-400M    & R-50         & \checkmark & 17.2      & \gray{27.0}\\
PromptDet~\cite{feng2022promptdet}    & ViT-B/32 & CLIP-400M & R-50  & \checkmark    & 21.4      & \gray{25.3}\\
OADB~\cite{OADP}   & ViT-B/32 & CLIP-400M & R-50 & \checkmark & {21.7}  & \gray{26.6}\\
RegionCLIP~\cite{zhong2021regionclip}       & R-50x4    & CLIP-400M + CC3M    & R-50x4       & \checkmark  & 22.0      & \gray{32.3} \\
CORA~\cite{wu2023cora}        & R-50x4   & CLIP-400M     & R-50x4     & \checkmark &{22.2}\textsuperscript{box} & - \\
Detic-CN2~\cite{zhou2022detecting}           & ViT-B/32 & CLIP-400M + INet-21K     & R-50       & WS    & 24.6      & \gray{32.4}\\
ViLD-Ens~\cite{gu2022openvocabulary}            & EffNet-B7 & ALIGN-1.8B    & EffNet-B7  & -   & 26.3      & \gray{29.3}\\
F-VLM~\cite{kuo2022fvlm}            & R-50x64 & CLIP-400M   & R-50x64  & -    & 32.8      & \gray{34.9} \\
\midrule
\bf\textit{ViT based:} & & & & \\
\gray{OWL-ViT~\cite{minderer2022simple}\textsuperscript{O365+VG}}              & \gray{ViT-L/14}  & \gray{CLIP-400M}    & \gray{ViT-L/14}  & \gray{-}    & \gray{25.6\textsuperscript{box}}     & \gray{34.7\textsuperscript{box}} \\
\gray{OWL-ViT v2~\cite{minderer2023scaling}\textsuperscript{O365+VG}}              & \gray{ViT-L/14}  & \gray{WebLI-10B}    & \gray{ViT-L/14}  & \gray{\checkmark}    & \gray{45.9\textsuperscript{box}}     & \gray{50.4\textsuperscript{box}} \\

{RO-ViT~\cite{rovit}}    & ViT-B/16 & ALIGN-1.8B    & ViT-B/16   & -    & {28.0} & \gray{30.2}\\
{RO-ViT~\cite{rovit}} $\dagger$   & ViT-L/16  & LAION-2B    & ViT-L/16   & -    & {32.4} & \gray{32.9} \\
CFM-ViT~\cite{kim2023contrastive}  & ViT-B/16 & ALIGN-1.8B     & ViT-B/16  & -     & {28.8} & \gray{32.0} \\
CFM-ViT~\cite{kim2023contrastive}  & ViT-L/16  & ALIGN-1.8B    & ViT-L/16   & -    & {33.9} & \gray{36.6} \\
CFM-ViT~\cite{kim2023contrastive} * & ViT-L/16  & LAION-2B    & ViT-L/16   & -    & {33.8} & \gray{36.4} \\
\bf{\ours (ours)}  & ViT-{{S}}/16  & LAION-2B     & ViT-{{S}}/16   & -    & \bf{26.2} & \gray{28.8} \\
\bf{\ours (ours)} & ViT-{{B}}/16 & LAION-2B   & ViT-{{B}}/16    & -   & \bf{31.5} & \gray{32.4} \\
\bf{\ours (ours)} & ViT-{{L}}/16 & LAION-2B      & ViT-{{L}}/16   & -    & \bf{37.6} & \gray{36.2} \\
\bf{\ours (ours)} & ViT-{{L}}/16  & DataComp-1B    & ViT-{{L}}/16   & -    & \bf{40.5} & \gray{38.0} \\
\bottomrule
\end{tabular} 
}
\vspace{1mm}
\caption{\small{\textbf{LVIS open-vocabulary detection (mask AP).} \ours outperforms the best existing approach by +3.7 mask AP$_r$. WS: uses weak supervision from ImageNet-21K. $\dagger$: reports LAION-2B results in arXiv version. *: our reproduced result using LAION-2B. \gray{O365+VG}: uses extra Objects365 and Visual Genome data.}}
\label{tab:ovd_lvis}
\vspace{-6mm}
\end{table}

\begin{table}[t]
\centering
\tablestyle{3.0pt}{0.9}
\scriptsize{
\begin{tabular}{lcccccc}
\toprule
method & \makecell[c]{pretraining \\ model} & \makecell[c]{pretraining \\ data} & \makecell[c]{detector \\ backbone }  & \makecell[c]{w/ pseudo \\ box labels}  & \bf{novel AP} &  \gray{AP}\\
\midrule
\bf\textit{ConvNet based:} & & & & & & \\
OVR-CNN~\cite{Zareian_2021_CVPR}  & R-50  & CLIP-400M + COCO-Cap & R-50 & - & 22.8 & 39.9 \\
ViLD~\cite{gu2022openvocabulary} & ViT-B/32 & CLIP-400M & R-50 & -  & 27.6  & \gray{51.3} \\
F-VLM~\cite{kuo2022fvlm}         & R-50  & CLIP-400M & R-50  & -   & 28.0 & \gray{39.6} \\
OV-DETR~\cite{zang2022open}      & ViT-B/32 & CLIP-400M & R-50  & -   & 29.4  & \gray{52.7} \\
CoDet~\cite{ma2024codet}  & R-50  & CLIP-400M + COCO-Cap  & R-50  & -  & 30.6  & \gray{46.6} \\
PromptDet~\cite{feng2022promptdet}          & ViT-B/32 & CLIP-400M & R-50  & \checkmark   & 26.6  & \gray{50.6} \\
XPM~\cite{huynh2022open}        & R-50 & CLIP-400M  & R-50  & \checkmark   & 27.0   & \gray{41.2} \\
OADB~\cite{OADP}   & ViT-B/32 & CLIP-400M + COCO-Cap & R-50 & \checkmark & {30.0}  & \gray{47.2} \\
VL-PLM~\cite{zhao2022exploiting}            & ViT-B/32 & CLIP-400M  & R-50  & \checkmark & 34.4  & \gray{53.5} \\
RegionCLIP~\cite{zhong2021regionclip} & R-50x4 & \makecell[c]{CLIP-400M + CC3M \\ + COCO-Cap}  & R-50x4 & \checkmark & 39.3  & \gray{55.7} \\
EdaDet~\cite{shi2023edadet}  & R-50  & CLIP-400M   & R-50  & \checkmark   & 40.2  & \gray{52.5} \\
CORA~\cite{wu2023cora}  & R-50x4  & CLIP-400M   & R-50x4  & \checkmark   & 41.7  & \gray{43.8} \\
Detic-CN2~\cite{zhou2022detecting}          & ViT-B/32  & CLIP-400M + INet-21K  & R-50      & WS    & 27.8      & \gray{45.0} \\

\midrule
\bf\textit{ViT based:} & & & & & & \\
{RO-ViT}~\cite{rovit}    & ViT-B/16 & ALIGN-1.8B   & ViT-B/16     & -  & {30.2} & \gray{41.5} \\
RO-ViT~\cite{rovit}    & ViT-L/16 & ALIGN-1.8B   & ViT-L/16    & -    &  {33.0} & \gray{47.7} \\
RO-ViT~\cite{rovit} *    & ViT-L/16 & LAION-2B   & ViT-L/16    & -    &  {33.3} & \gray{47.9} \\
CFM-ViT~\cite{kim2023contrastive}    & ViT-B/16 & ALIGN-1.8B   & ViT-B/16    & -    &  {30.8} & \gray{42.4} \\
CFM-ViT~\cite{kim2023contrastive}    & ViT-L/16 & ALIGN-1.8B  & ViT-L/16     & -   &  {34.1} & \gray{46.0} \\
CFM-ViT~\cite{kim2023contrastive} *   & ViT-L/16 & LAION-2B  & ViT-L/16     & -   &  {34.3} & \gray{46.4} \\
\bf{\ours (ours)}   & ViT-S/16 & LAION-2B  & ViT-S/16     & -   &  \bf{32.3} & \gray{44.4} \\
\bf{\ours (ours)}   & ViT-B/16 & LAION-2B  & ViT-B/16     & -   &  \bf{36.6} & \gray{48.8} \\
\bf{\ours (ours)}    & ViT-L/16 & LAION-2B & ViT-L/16     & -   &  \bf{39.6} & \gray{54.4} \\ 
\bf{\ours (ours)}    & ViT-L/16 & DataComp-1B & ViT-L/16     & -   &  \bf{40.2} & \gray{54.6}\\ 
\bottomrule
\end{tabular}
}
\vspace{1mm}
\caption{\small\textbf{COCO open-vocabulary detection (box AP50).} \ours demonstrates a very competitive novel category AP without using pseudo labeling or weak supervision (WS). *: our reproduced result using LAION-2B.
}
\vspace{-6mm}
\label{tab:ovd_coco}
\end{table}

\begin{table}[t]
\centering
\tablestyle{8pt}{0.9}
\scriptsize{
\begin{tabular}{llccc}
\toprule
method              & backbone  & AP    & AP\textsubscript{50}  & AP\textsubscript{75}    \\
\midrule
\gray{Supervised~\cite{gu2022openvocabulary}} & \gray{R-50} & \gray{25.6} & \gray{38.6} & \gray{28.0}  \\
ViLD~\cite{gu2022openvocabulary}   & R-50      & 11.8      & 18.2      & 12.6  \\
DetPro~\cite{du2022learning}       & R-50      & 12.1      & 18.8      & 12.9  \\
BARON~\cite{wu2023aligning}       & R-50      & 13.6       & 21.0      & 14.5  \\
F-VLM~\cite{kuo2022fvlm} & R-50x16 & 16.2 & 25.3 & 17.5 \\
F-VLM~\cite{kuo2022fvlm} & R-50x64 & 17.7 & 27.4 & 19.1 \\
\midrule
RO-ViT~\cite{rovit}               & ViT-L/16  & {17.1}    & {26.9}    & {18.5} \\
CFM-ViT~\cite{kim2023contrastive}  & ViT-L/16  & {18.7}    & {28.9}    & {20.3} \\
\bf{\ours (ours)}   & ViT-L/16  & \bf{20.0}    & \bf{31.8}    & \bf{21.5} \\
\bottomrule
\end{tabular}
}
\vspace{1mm}
\caption{\small\textbf{Zero-shot transfer detection from LVIS$_{base}$ to Objects365 (box AP).} All models are tested on Objects365 dataset following the setup of~\cite{gu2022openvocabulary}.}
\vspace{-6mm}
\label{tab:transfer}
\end{table}

\vspace{-2mm}
\subsection{Main Results}
\vspace{-2mm}

\paragraph{LVIS Benchmark.}\quad
In~\tabref{tab:ovd_lvis}, we report the comparison with existing methods on the challenging LVIS benchmark. The `frequent' and `common' classes of the dataset belong to the base categories C$_B$, and the `rare' classes are the novel categories C$_N$. 
The primary metric is the mask AP on rare classes (mask AP$_r$). The \ours model achieves the performance of 37.6 mask AP$_r$, which significantly outperforms the state-of-the-art approach RO-ViT~\cite{rovit} with the same ViT-L backbone by +5.1 points using the same pretraining data LAION-2B~\cite{schuhmann2021laion}. We also outperform the state-of-the-art CFM-ViT~\cite{kim2023contrastive} by +3.6 points.
Our best performance sets a new state-of-the-art 40.9 mask AP$_r$ when using DataComp-1B~\cite{gadre2023datacomp} in pretraining.
With the ViT-B backbone, \ours maintains a healthy margin of around +2.5 AP$_r$ above existing ViT-B based approaches.

\paragraph{COCO Benchmark.}\quad
We present the comparison on COCO benchmark in \tabref{tab:ovd_coco}. The main metric is AP50 of novel categories (novel AP). Without using pseudo labeling~\cite{feng2022promptdet, huynh2022open,zhao2022exploiting,zhong2021regionclip}, weak supervision~\cite{zhou2022detecting}, or externally trained detector modules~\cite{rasheed2022bridging, wu2023aligning}, our model demonstrates competitive results of 39.6 novel AP with LAION-2B and 40.2 with DataComp-1B. Among the ViT-based methods, \ours outperforms recent works RO-ViT~\cite{rovit} and CFM-ViT~\cite{kim2023contrastive} by a clear margin of +6.3 and +5.3 points, respectively.

\begin{table*}[t]
\centering
\subfloat[\scriptsize{\textbf{\ours framework}.\label{tab:ablation:overall}}]{
\tablestyle{2.0pt}{1.0}
\scriptsize{\begin{tabular}{lcc}
{method}  & {AP$_r$}  &   \gray{AP} \\
\toprule
RO-ViT~\cite{rovit}        & 32.4 {\white{(+0.0)}}    & \gray{32.9}  \\
\arrayrulecolor{gray}\hline
RO-ViT \textit{our repro.}   & 32.2 {\white{(+0.0)}}    & \gray{33.0}  \\
w/ REP                   & 34.8 ({\blue{+2.6}})     & \gray{34.9}  \\
w/ SWL                   & 35.0 ({\blue{+2.8}})     & \gray{35.2} \\
w/ REP +SWL             & 36.3 ({\blue{+4.1}})     & \gray{35.8}  \\
\baseline{w/ REP +SWL +FD}    & \baseline{37.6 ({\blue{+5.4}})}     & \baseline{\gray{36.2}}  \\
\end{tabular}}} \hspace{1mm}
\subfloat[\scriptsize{\textbf{Detector components in pretraining}.\label{tab:ablation:dito_pretraining}}]{
\tablestyle{1.9pt}{1.0}
\scriptsize{\begin{tabular}{lcc}
{pretraining method}  & {AP$_r$}  &   \gray{AP} \\
\toprule
RO-ViT \textit{our repro.} & 32.2 {\white{(+0.0)}}    & \gray{33.0}  \\
w/ FPN                   & 33.2 ({\blue{+1.0}})     & \gray{33.7}  \\
{w/ FPN +Head}       & {34.2 ({\blue{+2.0}})}     & {\gray{34.2}} \\
\baseline{w/ FPN +Head +RPN} & \baseline{34.8 ({\blue{+2.6}})}   & \baseline{\gray{34.9}}
\\ \\ \\
\end{tabular}}}\hspace{0.9mm}
\subfloat[\scriptsize{\textbf{RoI sampling and pooling.} \label{tab:ablation:dito_roi}}]{
\tablestyle{1.8pt}{1.0}
\scriptsize{\begin{tabular}{lcc}
{RoI embedding}  & AP$_r$  &   \gray{AP} \\
\toprule
global avg / lvl               & 34.0     & \gray{33.9}  \\
global max / lvl               & 33.7     & \gray{33.5}  \\
multi RoIs, avg / lvl & 33.8     & \gray{34.0}  \\
\baseline{multi RoIs, max / lvl}  & \baseline{34.8}     & \baseline{\gray{34.9}}  \\
multi RoIs, max all  & 34.1     & \gray{34.3}  \\
\\
\end{tabular}}}
\vspace{-1mm}
\caption{\small\textbf{Ablation on overall \ours framework and \Detection Pretraining (\secref{sec:rep}).} REP: Region-centric Pretraining, SWL: Shifted-Window Learning, FD: Frozen backbone Distillation. Best setting is in \hl{gray}.}
\vspace{-5mm}
\label{tab:dito_ablations}
\end{table*}

\begin{table*}[t]
\centering
\subfloat[\scriptsize{\textbf{Shifted-window learning}.\label{tab:ablation:swl1}}]{
\tablestyle{6.0pt}{1.0}
\scriptsize{
\begin{tabular}{lcc}
{backbone}  & {AP$_r$} & \gray{AP}\\
\toprule
fully global attn.           & {33.4} &  \gray{33.8} \\
baseline window attn.~\cite{rovit}     & {32.2} &  \gray{33.0} \\
Swin~\cite{swin} style   & 31.3   &  \gray{33.1} \\
\baseline{shifted window}           & \baseline{35.0} &  \baseline{\gray{35.2}} \\
\end{tabular}}}
\hspace{5mm}
\subfloat[\scriptsize{\textbf{Effect of SWL w.r.t. \# global attention layers}.\label{tab:ablation:swl2}}]{
\tablestyle{8.0pt}{1.0}
\scriptsize{
\begin{tabular}{ccc}
{\# global attn. layer}     & {base}         & {w/ SWL} \\
\toprule
0               &    {30.7}     & {34.6} ({\blue{+3.9}}) \\
4               &    {32.2}     & \baseline{35.0} ({\blue{+2.8}}) \\
12              &    32.4       & 34.2 ({\blue{+1.8}}) \\
24 (all layers) &    {33.4}     & {33.4} ({\blue{+0.0}}) \\
\end{tabular}}}
\vspace{-1mm}
\caption{\small\textbf{Ablation on Shifted-Window Learning (SWL - \secref{sec:swl}).} Best setting is in \hl{gray}.}
\vspace{-7mm}
\label{tab:swl}
\end{table*}

\begin{table*}[t]
\centering
\subfloat[\scriptsize{\textbf{Frozen backbone distillation (GT region recognition)}.\label{tab:ablation:dist}}]{
\tablestyle{3.0pt}{1.0}
\scriptsize{
\begin{tabular}{ccccc}
{backbone}     & AP & AP$_r$ & AP$_c$ & AP$_f$ \\
\toprule
\multicolumn{5}{l}{\textit{\textbf{GT boxes given (region classification):}}} \\
before detection training               & 48.7  & 56.8 & 51.5 & 42.1 \\
after detection training                & {53.7} & 54.2 & 53.7 & 53.5 \\
after detection training w/ FD     & {54.8} & 57.7 & 54.3 & 53.2  \\
\end{tabular}}}
\hspace{3mm}
\subfloat[\scriptsize{\textbf{Effect of frozen backbone distillation}.\label{tab:ablation:dist_2}}]{
\tablestyle{4.0pt}{1.0}
\scriptsize{\begin{tabular}{lcc}
{method}  & {AP$_r$}  &   \gray{AP} \\
\toprule
RO-ViT \textit{our repro.}  & 32.2 {\white{(+0.0)}}    & \gray{33.0}  \\
RO-ViT \textit{our repro.} + FD  & 34.0 {({\blue{+1.8}})}    & \gray{33.6}  \\
\arrayrulecolor{gray}\hline
w/ REP + SWL       & 36.3 {\white{(+0.0)}}      & \gray{35.8}  \\
\baseline{w/ REP + SWL + FD}    & \baseline{37.6 ({\blue{+1.3}})}  & \baseline{\gray{36.2}} \\
\end{tabular}}}
\vspace{-1mm}
\caption{\small\textbf{Ablation on frozen backbone distillation (FD - \secref{sec:dist}).} Best setting is in \hl{gray}.}
\vspace{-6mm}
\label{tab:overall}
\end{table*}

\paragraph{Transfer detection.}\quad
We further evaluate \ours in the transfer detection setting, where the open-vocabulary detector trained on one dataset (LVIS$_{base}$) is tested on another dataset (Objects365) without any finetuning. By simply replacing the text embeddings, \tabref{tab:transfer} shows that \ours achieves  20.0 AP, outperforming previous methods using ConvNet or ViT backbones of similar size.

\vspace{-1mm}
\subsection{Ablation Studies}
\vspace{-2mm}
For ablation study, we use the ViT-L/16 model pretrained with LAION-2B, and evaluate on the LVIS benchmark and report mask AP$_r$.

\paragraph{\ours overall framework.}\quad
\tabref{tab:ablation:overall} summarizes the benefits of each \ours components. The \detection pretraining (REP) improves the contrastive model baseline by +2.6 AP$_r$ and shifted window learning (SWL) improves the baseline by +2.8 points. Combining both strategies brings a significant gain of +4.1 AP$_r$.
{Lastly, we observe that incorporating the frozen backbone distillation leads to an additional boost of +1.2 points.} In the following, we provide ablations for each components.

\paragraph{\Detection Pretraining.}\quad
In \tabref{tab:ablation:dito_pretraining}, we ablate our \detection image-text pretraining by progressively adding the FPN, Faster R-CNN head, and RPN into the pretraining. Our `baseline' is the contrastive image-text pretraining with cropped positional embedding~\cite{rovit}. On top of this, `w/ FPN' introduces the FPN into the pretraining, where each pyramid level (whole image) map is mean-pooled into an image embedding, followed by the image-text contrastive loss per level. It improves the baseline by +1.0 AP$_r$. Adding the Faster R-CNN head further improves the alignment between the pretraining and detection finetuning, showing a gain of +2.0 AP$_r$. Incorporating all components \ie the FPN, Faster R-CNN and RPN heads achieves the best 34.8 AP$_r$, a significant gain of +2.6 points over the baseline.

\tabref{tab:ablation:dito_roi}  shows that both global avg- and max-pooling are sub-optimal due to the lack of saliency map (avg-pool), and the limited capacity of a single pixel to represent semantic concepts for contrastive learning (max-pool), respectively. Our approach combines the best of both worlds, by first avg-pooling within each RoI, and then max-pooling over these RoI embeddings. Each embedding represents a proper RoI and the global representation captures the saliency map through max-pooling. Pooling per pyramid level (\ie multi-level image-text supervision) outperforms pooling over all levels.

\begin{figure*}[]
\centering
\vspace{-5mm}
\includegraphics[width=1.0\linewidth]{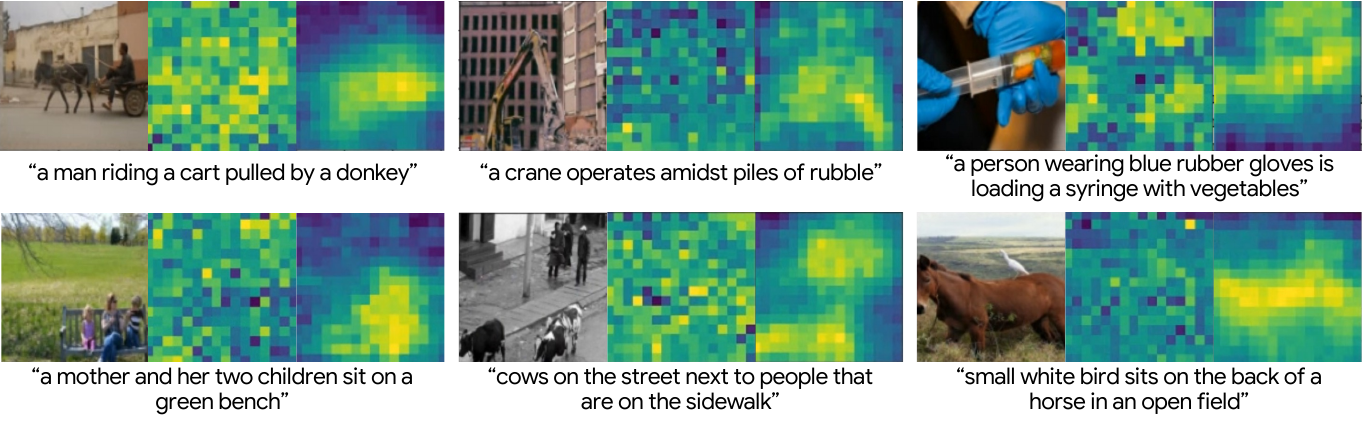}
\vspace{-7mm}
\caption{\small\textbf{Visual-text similarity map}. For each example, we show the paired image (left) and text (bottom) input, and the visual-text similarity map using the contrastive model baseline~\cite{rovit} backbone features (middle) or our \detection pretraining features (right). We use Flickr30K (top row) and COCO Captions (bottom row) datasets.}
\label{fig:heatmap}
\vspace{-6mm}
\end{figure*}

\paragraph{Shifted-Window Learning for detection.}\quad
The CLIP ViT backbones are initially pretrained on lower resolutions and then adapted to higher resolution detection images. 
In \tabref{tab:ablation:swl1}, we assess the efficacy of the shifted-window backbone in open-vocabulary detection training. 
Although the fully global attention model improves the detection task compared to the baseline windowed attention model~\cite{rovit}, directly applying the pretrained ViT on high-resolution inputs may not be optimal, potentially compromising the locality structure of the pretrained lower-resolution features. Our shifted-window learning approach achieves 35.0 AP$_r$ by preserving the locality structure from windowed attention as well as integrating information across fixed windows, outperforming the fully global attention model (33.4 AP$_r$) which is computationally intensive. Notably, OWL-ViT~\cite{minderer2022simple, minderer2023scaling} adopts the fully global attention model.
In addition, naively applying the layer-alternating shifted window as in Swin~\cite{swin} leads to a performance drop (see \secref{sec:swl}).
\tabref{tab:ablation:swl2} delves deeper into the behavior of SWL. The advantage of SWL diminishes steadily with increasing number of global attention layers in the windowed attention backbone, validating its better information mixing enabled by the SWL.

\paragraph{Frozen backbone distillation.}\quad
\tabref{tab:ablation:dist} studies the region classification capability of the CLIP ViT backbone before and after the detection finetuning. We use the ground truth boxes and measures AP scores to evaluate the zero-shot region classification of the base (`frequent' + `common') and novel (`rare') categories. We observe that the finetuned backbone indeed shows a notable drop of -2.6 AP$_r$ on novel classes, while overfitting to the {base} classes.
The frozen backbone distillation (`w/ FD') leads to a significant improvement of +3.5 AP$_r$ even surpassing the frozen pretrained backbone (before detection finetuning), while maintaining performance on the base classes (AP$_c$ and AP$_f$). 
\tabref{tab:ablation:dist_2} presents the open-vocabulary detection results where the frozen backbone distillation brings a clear gain of +1.3$\sim$1.8 AP$_r$.
These results highlight the efficacy of our frozen backbone distillation in detection finetuning (\secref{sec:dist}), as it effectively preserves the pretrained open-vocabulary knowledge while acquiring explicit region-text alignment through detection supervision.

\vspace{-2mm}
\subsection{Visualization}
\vspace{-2mm}
In~\figref{fig:heatmap}, we visualize the similarity map between the image features and a query text embedding using the Flickr30K~\cite{plummer2015flickr30k} and COCO Captions~\cite{chen2015microsoft} datasets.
For each sample, we compare the baseline contrastive model~\cite{rovit} backbone features (middle) and \detection pretrained features (right). We select pyramid level 4 which has the same resolution as the backbone features and apply the Faster R-CNN head in a sliding window manner to obtain the dense feature map.
We observe that \detection pretraining captures more localized semantic information on the image-text pairs. 

In~\figref{fig:teaser}, we visualize the \ours outputs on LVIS novel categories and Ego4D~\cite{grauman2022ego4d} which is real-world and out-of-distribution data. We use the same \ours detector trained on LVIS$_{base}$. The categories for Ego4D are provided by the user based on visual inspection of the video. We observe that \ours is able to capture many novel and unseen objects even under the significant domain shift.

%% file: sections/5_conclusion.tex
\vspace{-2mm}
\section{Conclusion}
\vspace{-2mm}
We introduce DITO, a region-centric approach for open-vocabulary detection using large-scale image-text pairs. By integrating detection architecture onto the image backbone in CLIP pretraining, it learns locality-sensitive information without requiring pseudo labeling or box annotations. Furthermore, we propose a shifted-window learning method to mitigate the bias of the window attention pattern in CLIP ViT detectors.
Experiments show that \ours outperforms the state-of-the-art by large margins on the LVIS benchmark, and is very competitive on the COCO benchmark and transfer detection. We hope this work would inspire the community to explore \detection image-language pretraining for open-vocabulary localization tasks.

%% file: sections/6_appendix.tex
\appendix
\section{Additional Implementation Details}
\paragraph{Region-centric Pretraining.}\quad
As mentioned in Sec 3.2, our Region-centric Pretraining (REP) employs the multi-level image-text supervision and RPN-objectness training. The multi-level image-text supervision consists in the standard image-text contrastive loss ($L_{con}$) applied at each $i$-th feature pyramid level. 
During training, we employ the multi-level visual-text similarity as a supervisory signal for training the RPN's objectness map (see Fig. 3). The target RPN score is computed as the cosine similarity between each RoI embeddings and the text, where any negative dot product value is mapped to zero to keep the target score in range [0, 1]. We use L1 regression loss between the target scores and the corresponding RoIs' center locations on the objectness map.
In sum, the total loss objectives of our region-centric pretraining is $L_{REP} = \sum_{i=2}^{5}L_{con}^{i} + \lambda L_{reg}$, where $\lambda=1$.
\tabref{tab:hparams_pt} summarizes the hyperparameters used in our \Detection Pretraining.

\begin{table}
\centering
\tablestyle{10.0pt}{1.0}
\footnotesize{
\begin{tabular}{lcc}
 & baseline CLIP (Sec {\red{3.1}}) & {\Detection Pretraining (Sec {\red{3.2}})}  \\
\toprule
optimizer           & AdamW         & AdamW         \\
momentum            & $\beta$=0.9   & $\beta$=0.9   \\
weight decay        & 0.01          & 0.01          \\
learning rate       & 0.001         & 0.0001        \\
warmup steps        & 5k            & 5k            \\
total steps         & 500k          & 30k           \\
batch size          & 16384         & 4096          \\
image size          & 224           & 256           \\  
\end{tabular}
\vspace{1mm}
\caption{\small{Hyperparameters for \Detection Pretraining.} }
\label{tab:hparams_pt}
}
\end{table}

\paragraph{Open-vocabulary detection finetuning.}\quad
We follow the same objective functions of Mask R-CNN (for LVIS) and Faster R-CNN (for COCO), except that we have an additional frozen backbone distillation loss (Sec. 3.4). 
We use a cosine distance loss that aligns the RoI-Align embeddings extracted from the feature maps of finetuned vs frozen backbones. The cosine distance is computed for each RoI then averaged over all RoIs.
In sum, our detection loss objectives is $L_{Det} = L_{Rpn\text{-}obj} + L_{Rpn\text{-}box} + L_{Frcnn\text{-}class} + L_{Frcnn\text{-}box} + L_{mask} + \gamma L_{distill}$, where $\gamma = 1$.

\tabref{tab:hparams_ft} summarizes the hyperparameters used in our open-vocabulary detection finetuning. We use the same open-vocabulary detector design of RO-ViT~\cite{rovit} which adopts the ViTDet architecture~\cite{li2022exploring} and the centerness-based RPN~\cite{kim2022learning} that uses a single anchor per location.

\begin{table}
\centering
\tablestyle{10.0pt}{1.0}
\begin{tabular}{lcc}
OVD finetuning  &   ViT-L (LVIS / COCO) &   ViT-B and S (LVIS / COCO) \\
\toprule
optimizer           & SGD            & SGD           \\
momentum            & $\beta$=0.9    & $\beta$=0.9   \\
weight decay        & 0.0001         & 0.0001         \\
learning rate       & 0.18 / 0.02    & 0.36 / 0.02   \\
backbone lr ratio   & 0.6$\times$ / 0.2$\times$  & 0.1$\times$ / 0.1$\times$  \\
step decay factor   & 0.1$\times$   & 0.1$\times$   \\
step decay schedule & [0.8, 0.9, 0.95] & [0.8, 0.9, 0.95] \\
warmup steps        & 1k            & 1k            \\
total steps         & 36.8k / 11.3k    & 46.1k / 11.3k    \\
batch size          & 128           & 256           \\
image size          & 1024          & 1024          \\
\end{tabular}
\vspace{1mm}
\caption{\small{Hyperparameters for open-vocabulary detection finetuning.} }
\label{tab:hparams_ft}
\end{table}

\section{Additional Ablations}
\paragraph{Region-centric Pretraining (REP).} \quad
\tabref{tab:ablation:dito_roi} provides more ablations on the RoI sampling and pooling methods in the FPN within the REP training. We investigate whether pooling over random boxes are more effective than global pooling over pixels or blockwise pooling on a regular grid. 
\tabref{tab:ablation:dito_roi}  shows that both global avg- and max-pooling are sub-optimal due to the lack of saliency map (avg-pool), and the limited capacity of a single pixel to represent semantic concepts for contrastive learning (max-pool), respectively. Our approach combines the best of both worlds, by first avg-pooling within each RoI, and then max-pooling over these RoI embeddings. Each embedding represents a proper RoI and the global representation captures the saliency map through max-pooling. Despite the absence of explicit region-level supervision in our REP training, the max pooling over random regions encourages the more salient, text-aligned RoI features to contribute more to the whole image representation in the contrastive loss. To study the need of randomness in our method, we divide the feature map into a $N \times N$ grid and treat each grid cell as an RoI (block-wise RoI). The random RoI is superior due to the greater variety in RoI scales and locations.

\tabref{tab:ablation:dito_batch} ablates contrastive batch size in our REP pretraining, where we choose batch size 4k, as larger batch does not result in improvements.

\begin{table*}[t]
\centering
\subfloat[\small{\textbf{RoI sampling and pooling:} The pooling is applied per pyramid level for all methods. Using multiple random RoIs followed by max pooling performs the best, outperforming the whole-image RoI and pixel-wise RoIs methods. \label{tab:ablation:dito_roi}}]{

\tablestyle{8.0pt}{1.0}
\begin{tabular}{lccc}
RoI sampling & pool  & AP$_r$  &   \gray{AP} \\
\toprule
global (pixel-wise RoI)  & avg   & 34.0     & \gray{33.9}  \\
global (pixel-wise RoI)  & max   & 33.7     & \gray{33.5}  \\
block-wise (8$\times$8 grid RoI) & max       & 33.9     & \gray{33.8}  \\
block-wise (4$\times$4 grid RoI) & max       & 34.2    & \gray{34.3}  \\
block-wise (2$\times$2 grid RoI) & max       & 33.1     & \gray{33.8}  \\
\baseline{random RoI} & \baseline{max}  & \baseline{34.8}     & \baseline{\gray{34.9}}  \\ 
\end{tabular}
}
\hspace{7mm}
\subfloat[\small{\textbf{Contrastive batch size:} We choose batch size 4k, as larger batch doe not show improvement.\label{tab:ablation:dito_batch}}]{
\tablestyle{7pt}{1.0}
{\begin{tabular}{ccc}
{batch}  & AP$_r$ &   \gray{AP} \\
\toprule
1k      & {33.9}     & \gray{34.1}  \\
2k      & {34.3}     & \gray{34.6}  \\
\baseline{4k}      & \baseline{34.8}     & \baseline{\gray{34.9}}  \\
16k     & {34.7}     & \gray{34.8}  \\ \\ \\ 
\end{tabular}}}
\caption{\small{More ablations for Region-centric Pretraining (Sec. 3.1).} (a) RoI sampling and pooling in the FPN. (b) Contrastive batch size for our region-centric pretraining. Best setting is in \hl{gray}.}
\vspace{0mm}
\label{tab:dito_ablations}
\end{table*}

\paragraph{Shifted-Window Learning for Detection (SWL).}\quad
The proposed SWL is beneficial for both OVD and fully supervised detection. In ~\tabref{tab:ablation:swl1}, we show that the gain from SWL is 50\% larger for \textit{rare} classes (+2.8 AP) than {frequent} and {common} classes (+1.9 AP) in the LVIS OVD benchmark. For standard detection, Tab.~\ref{tab:ablation:swl2} shows SWL improves performance using both pretrained or randomly initialized backbone. The results show that SWL can improve detection in general, as well.

\begin{table}[t]
\centering
\subfloat[{\small{LVIS OVD benchmark}.\label{tab:ablation:swl1}}]{
\tablestyle{4.0pt}{1.0}
\begin{tabular}{lccc}
{backbone}      & {mask AP$_r$}  & {mask AP$_c$} & {mask AP$_f$}  \\
\toprule
baseline win. attn.
                & 32.2      & 33.6      & 33.1 \\
\baseline{SWL (ours)}            & \baseline{35.0 (\blue{+2.8})}     & \baseline{35.5 (\blue{+1.9})}      & \baseline{34.9 (\blue{+1.8})} \\
\end{tabular}}
\vspace{1mm}
\subfloat[{\small{{{Fully-supervised}} detection on COCO (ViT-B)}. 
\label{tab:ablation:swl2}}]{
\tablestyle{4.0pt}{1.0}
\begin{tabular}{lcc}
{backbone}      & {AP (pretrained)} & AP (random init.) \\
\toprule
baseline win. attn.
                & 48.0              & 39.5 \\
\baseline{SWL (ours)}      & \baseline{49.0   (\blue{+1.0})}            & \baseline{40.3 (\blue{+0.8})}  \\
\end{tabular}}
\caption{{Generality of Shifted Window Learning (SWL)}}
\label{tab:swl}
\end{table}

\section{Limitations}
Our models utilize the rich image-text information acquired through pretraining, which may reinforce deficiencies and biases in the raw web data and expose potentially harmful biases or stereotypes. The models we trained are designed for academic research purposes and need more rigorous fairness studies or data cleaning before serving product applications.